\documentclass[sigconf]{aamas} 
\usepackage{balance} % for balancing columns on the final page
\usepackage{algorithm}
\usepackage{algpseudocode}

\setcopyright{ifaamas}
\acmConference[AAMAS '24]{Proc.\@ of the 23rd International Conference
on Autonomous Agents and Multiagent Systems (AAMAS 2024)}{May 6 -- 10, 2024}
{Auckland, New Zealand}{N.~Alechina, V.~Dignum, M.~Dastani, J.S.~Sichman (eds.)}
\copyrightyear{2024}
\acmYear{2024}
\acmDOI{}
\acmPrice{}
\acmISBN{}

\acmSubmissionID{<<EasyChair submission id>>}

\title[AAMAS-2024 Formatting Instructions]{Applying Multi-Agent Negotiation to Solve the Production Routing Problem With Privacy Preserving}

\author{Luiza Pellin Biasoto}
\affiliation{
  \institution{Escola Politécnica \\ Universidade de São Paulo (USP)}
  \city{São Paulo}
  \country{Brazil}}
\email{luizapb@usp.br}

\author{Vinicius Renan de Carvalho}
\affiliation{
  \institution{Escola Politécnica \\ Universidade de São Paulo (USP)}
  \city{São Paulo}
  \country{Brazil}}
\email{vrcarvalho@usp.br}

\author{Jaime Simão Sichman}
\affiliation{
  \institution{Escola Politécnica \\ Universidade de São Paulo (USP)}
  \city{São Paulo}
  \country{Brazil}}
\email{jaime.sichman@usp.br}

\begin{abstract}
This paper presents a novel approach to address the Production Routing Problem with Privacy Preserving (PRPPP) in supply chain optimization. The integrated optimization of production, inventory, distribution, and routing decisions in real-world industry applications poses several challenges, including increased complexity, discrepancies between planning and execution, and constraints on information sharing. To mitigate these challenges, this paper proposes the use of intelligent agent negotiation within a hybrid Multi-Agent System (MAS) integrated with optimization algorithms. The MAS facilitates communication and coordination among entities, encapsulates private information, and enables negotiation. This, along with optimization algorithms, makes it a compelling framework for establishing optimal solutions. The approach is supported by real-world applications and synergies between MAS and optimization methods, demonstrating its effectiveness in addressing complex supply chain optimization problems.
\end{abstract}


\newcommand{\BibTeX}{\rm B\kern-.05em{\sc i\kern-.025em b}\kern-.08em\TeX}

\begin{document}

%%% The following commands remove the headers in your paper. For final 
%%% papers will be inserted during the pagination process.

\pagestyle{fancy}
\fancyhead{}

%%% The next command prints the information defined in the preamble.

\maketitle 

%%%%%%%%%%%%%%%%%%%%%%%%%%%%%%%%%%%%%%%%%%%%%%%%%%%%%%%%%%%%%%%%%%%%%%%%

\section{Introduction}

% \vinnie{COMENTARIOS GERAIS: Me colocando no papel de revisor eu tenho 2 impressoes, uma eh de que deveria ter mais imagens, especialmente sobre a definicao de agentes e negociacao. Acho q duas imagens bem feitas ja melhoraria em muito.}
% \jaime{de acordo}

Supply chain decisions proceed sequentially, with steps that typically operate in isolation with fixed parameters set by adjacent ones. For instance, in the distribution of paper production, a supply team identifies a distribution center that needs a specific product quantity, requesting it from a designated factory. Subsequently, the factory initiates production to meet the predetermined demands. Usually aiming to optimize operational costs, they are framed as \textbf{combinatorial optimization problems}. Methods like Mixed-Integer Programming or heuristics are then employed for effective planning.

Though solving steps separately may yield optimal or suboptimal, yet effective, solutions, integrated planning offers financial benefits. Studies show operational cost reductions ranging from 3\% to 20\% \cite{Chandra1994}, and a systematic review estimates an 11.08\% \cite{Hrabec2022} cost reduction compared to sequential solutions.

Several challenges make the integrated optimization of these decisions difficult in real-world industry applications: a) \textbf{Increased complexity of decisions}: Unifying business rules makes problems too large for commercial solutions, increasing the complexity of added variables and constraints significantly; b) \textbf{Discrepancy between planning and execution}: Operational reality complexities, like unforeseen events or lack of confidence in the provided solution, may lead to deviations between execution and planning strategies, necessitating real-time replanning. When these replannings do not occur through an optimization method, the decisions made can be suboptimal, resulting in financial losses; c) \textbf{Constraints on information sharing}: Privacy and information protection play an important role in real-world applications. Organizational constraints and privacy-preserving may limit access to essential information for decision-making. This can result in suboptimal plans or increased operational-level replanning due to a lack of a faithful representation of reality.

The increased complexity of decisions resulting from the integration can be mitigated by applying advanced optimization methods, such as decompositions, meta-heuristics, matheuristics, and hybrid optimization algorithms.

An automated system can be implemented to tackle the discrepancy between planning and execution. This system reads real-time data as input, identifies deviations, and re-executes optimization algorithms to reconstruct the optimal plan for immediate adherence.

However, conventional optimization methods may not effectively address constraints on information sharing, such as privacy preservation or information protection between departments. If crucial information is withheld from the optimization algorithm, it cannot incorporate it into its search for the optimal solution, potentially leading to suboptimal outcomes.

This paper addresses the three aforementioned challenges, creating a hybrid Multi-Agent System (MAS) integrated with optimization algorithms to solve the Production Routing Problem with Privacy Preserving (PRPPP). In short, we foresee that agents representing different clients can propose alternative solutions using exclusively local data, therefore enhancing privacy-preserving.

%\subsection{Multi-Agent Systems (MAS) and Optimization}
% \jaime{Num short paper, evitar usar subseções, vc perde espaço. Eu colocaria estes 2 proximos paragrafos, de forma mais resumida, no icicio da próxima seção}

A MAS automates plan generation and decision negotiation among entities, allowing diverse forms of reasoning to collaborate for solutions. Intelligent agents integrate with optimization algorithms to address complex optimization problems by decomposing them into simpler subproblems. Additionally, agents encapsulate private information, enabling negotiation without revealing strategies yet incorporating them into final planning solutions. 

In \cite{nof2023}, applications and synergies between agents and automation are explored. Additionally in \cite{Moghaddam2016}, collaborative control mechanisms are introduced to address real-time optimization challenges in the planning and resource allocation of small to medium-sized enterprises.

Furthermore, an application case of intelligent agents is presented in \cite{Kazemi2009}, where they handle decisions related to production and transportation. Additionally, agents can be used to reformulate problems as Distributed Constraint Optimization Problems (DCOP) when the problem is so complex or requires information privacy that its integrated resolution becomes unfeasible \cite{Furukita2022}. Agents can also coordinate solutions from different meta-heuristics to address Production and Distribution Planning Problems (PDPP) \cite{Kazemi2017} and even negotiate among themselves to determine the best meta-heuristic for solving a specific multi-objective problem \cite{Vini2022}. Other synergies between Evolutionary Computation (EC) and MAS are discussed in \cite{Vini2019}. 
%\section{The Production Routing Problem with Privacy Preserving (PRPPP)}
\section{THE PRODUCTION ROUTING PROBLEM (PRP)}
% \jaime{Nesta secao vc apresenta o PRP ou o PRPPP? Creio que seja o primeiro, então sugiro alterar o titulo}
Back to the paper industry example, the supply chain can be segmented into four distinct steps, each involving specific decisions: a) \textbf{Production}: determining the choice of product, size and its timing for manufacturing; b) \textbf{Inventory}: deciding the optimal size and duration for a product to remain in warehouses; c) \textbf{Distribution}: assigning products to specific distribution centers and scheduling their arrival times; and d) \textbf{Routing}: identifying the most efficient routing option based on a distribution plan.

The Production Routing Problem (PRP) includes all four steps integrated within its decision-making framework. Classical mathematical models compose the PRP, such as the Vehicle Routing Problem (VRP) \cite{Dantzig1959}, a well-known NP-hard problem, and the Lot-Sizing Problem (LSP) \cite{Wagner2004}.

% \jaime{Não creio que esta figura 1 seja necessaria see falatar espaço, pode sair.}
% \begin{figure}[!ht]
%   \centering
%   \includegraphics[width=0.75\linewidth]{figures/PRPdecomposition.png}
%   \caption{Supply chain planning models \cite{Adulyasak2014b}}
%   \label{fig:PRPdecomposition}
%   \Description{Supply chain planning models \cite{Adulyasak2014b}}
% \end{figure}

\begin{figure}[!ht]
  \centering
  \includegraphics[width=0.65\linewidth]{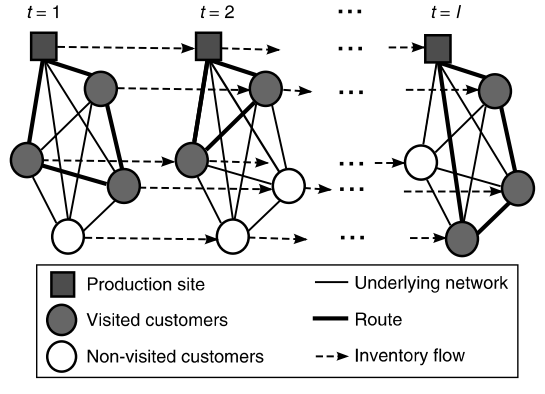}
  \caption{Production Routing Problem (PRP) \cite{Adulyasak2014b}}
  \label{fig:PRPflow}
  \Description{Production Routing Problem (PRP) \cite{Adulyasak2014b}}
\end{figure}

The domain of the PRP is defined by a complete graph \textit{G = (N, A)}, where \textit{N} represents the set of the supplier and retailers indexed by \textit{\(i \in \{0, ..., n\}\)} and \textit{\(A = \{(i,j): i, j \in N, i \neq j\}\)} is the set of arcs connecting the supplier and retailer. The supplier is represented by node 0, and the set of retailers is defined as \textit{\( N_{C} = N \textbackslash \{0\} \)}. A single product is manufactured in the factory and delivered to retailers by a set of identical vehicles \textit{\(K = \{1, ..., m\} \)} over a discrete and finite set of periods \textit{\(T = \{1, ..., l\}\)}, aiming to satisfy their demands in each period. Figure \ref{fig:PRPflow} depicts the graph \textit{G} with its nodes \textit{N} and arcs \textit{A}. 

The objective of the PRP is to provide the planning of deliveries and production for a determined time horizon while minimizing production costs, inventory costs (both at the supplier's and retailers' levels), and transportation costs:

\begin{equation} \label{fo}
min \sum_{t \in T}(up_{t} + fy_{t} + \sum_{i \in N}(h_{i}I_{it}) +\sum_{(i,j) \in A}(c_{ij}\sum_{k \in K}x_{ijkt} ))
\end{equation}

\textit{\(p_{t}\)} - production quantity in period \textit{t}; \textit{\(y_{t}\)} - equal to 1 if there is production at the factory in period \textit{t}, 0 otherwise; \textit{\(I_{it}\)} - inventory at node \textit{i} at the end of period \textit{t}; \textit{\(x_{ijkt}\)} - equal to 1 if a vehicle \textit{k} travels directly from node \textit{i} to node \textit{j} in period \textit{t} (see full PRP's formulation, decision variables and parameters description in Appendix).

A review of its mathematical formulations can be found in \cite{Adulyasak2015}. The PRP holds practical significance within a Vendor Managed Inventory (VMI) approach \cite{Adulyasak2015}. In this context, the supplier not only monitors retailers' inventory levels but also makes decisions regarding the replenishment policy for each retailer. 

It functions effectively when assuming the supplier has complete control over retailers' decisions. However, let's consider a scenario where the supplier lacks crucial information about retailers, such as their inventory costs, and retailers have the ability to negotiate with the supplier to expedite or defer certain deliveries. In this scenario, the PRP may no longer be entirely applicable, and the methods previously studied for its resolution may not be entirely suitable. The model that represents this specific scenario is the \textbf{Production Routing Problem with Privacy Preserving (PRPPP)}. Due to the privacy preservation, the term $\sum_{i \in N}(h_{i}I_{it})$ from the PRP's objective function, regarding the inventory costs of every node (i.e., supplier and retailers'), is affected and becomes \(h_{0}I_{0t}\), keeping only the inventory costs from the supplier, which they have access.

\section{The Production Routing Problem with Privacy Preserving (PRPPP)}
Assuming a PRPPP instance with a 6-month horizon, the solution output must contemplate the supplier's delivery plan to each retailer for the whole six months. Notably, not every retailer will receive deliveries every month; they may be concentrated in specific months to meet their demand. The variable transportation and production costs associated with each delivery period are reflected by the supplier in the form of shipping and product prices charged to retailers. In order to fulfill each retailer's demand plan with overall cost reduction, the supplier will receive the delivery preferences from retailers and propose optimal agendas for negotiation. These negotiations will be influenced by the changing shipping and product costs at each proposed delivery period, as well as the inventory costs unique to each retailer, which only they are aware of. The optimal negotiation agendas continue until a stopping criterion is met.

\subsection{Input Data}

The model takes as input data the parameters described in Section 2, including:

\begin{itemize}
\item Demand plan (\textit{\(d_{it}\)}): Each retailer (index \(i\)) must fulfill a specific demand plan; for example, requiring eight product units in months 3, 4, and 5 within a 6-month horizon  (index \(t\)),  totaling 24 units.

\item Supplier and retailers inventory costs (\textit{\(h_{i}\)}):  Distinct inventory costs influence retailer preferences when deciding on delivery negotiations. Supplier inventory costs are translated into product prices charged to retailers. These costs are fixed and known only to the respective supplier or retailer.

\item Unit production cost (\textit{u}) and setup cost (\textit{f}): These impact the supplier's production expenses, translated into product prices charged to retailers. Setup costs are fixed charges during production. Unit production costs fluctuate with the quantity produced; e.g., if the supplier produces 140 units in month 2 and 200 units in month 4, with variable production costs of 8 and setup costs of 1500, the total cost for the horizon is (140 * 8 + 1500) + (200 * 8 + 1500) = 5720.

\item Coordinates of the supplier and retailers: Geographical locations used to generate routes and compute transportation costs.

\item Transportation costs (\textit{\(c_{ij}\)}):  Calculated proportional to the Euclidean distance between the supplier and a retailer or between retailers. The total cost of a route is the sum of costs for each segment.

\item Maximum capacities: Maximum production (\textit{C}), vehicle (\textit{Q}) and inventory (\textit{\(L_{i}\)}) capacities may limit overall decisions.

\item Initial inventory levels (\textit{\(I_{i0}\)}): Initial levels for both suppliers and retailers at the planning horizon's start.

\end{itemize}

\subsection{Agents}

As seen in Figure \ref{fig:PRPflow}, the suppliers and retailers are the two agent types.

a) Supplier agent (Node 0 in set \textit{N}): aims to determine deliveries for each period to retailers while minimizing production and transportation costs. The supplier agent is a \textbf{coordinator agent}, responsible for coordinating retailer preferences, proposing optimal negotiation agendas, and mediating negotiations. The coordinator has total system knowledge, except for specific retailers' inventory costs. Their actions include:

\begin{itemize}
\item \textit{initialsol}: Generates an initial solution based on input data and retailers' ordered delivery preferences, considering the first delivery preference of each retailer while respecting production, vehicle and inventory capacities constraints.
\item \textit{optagenda}: Generates an optimal negotiation agenda, proposing \textbf{insertions, removals and substitutions} of retailers' deliveries.
\end{itemize}

b) Retailer agents (\textit{\( N_{C} = N \textbackslash \{0\} \)}): they have a demand plan and communicate delivery preferences to the supplier. Represented by \textbf{agents with partial knowledge}, they are aware of specific inventory costs but lack information about shipping and product prices until proposed during negotiations. Retailers know which retailers are part of their neighborhoods and their proximity. Their actions include:

\begin{itemize}
\item \textit{negotiate}: Participate in a negotiation agenda transaction, deciding on the suggested change based on their delta utility, i.e., selecting the change that would most satisfy the agent's preferences or goals.
\item \textit{vote}: Changes in neighborhood delivery plans may impact shipping and product prices from their retailers. Retailers are then selected to vote on the proposed changes; if the majority agrees, changes are implemented.
\end{itemize}

\subsection{Other Components}

\begin{itemize}
\item Neighborhood: Retailers access information about their neighborhood, i.e., other retailers with deliveries in the same period and corresponding Euclidean distances. The supplier updates the neighborhood with any solution plan changes.

\item Planning board: Carries initial and current solutions, updated after each successful negotiation. The supplier has access to all information, while retailers only see what directly affects them.

\item Transaction pool: Contains information about the current transaction, including negotiating retailers, affected retailers selected to vote on proposed changes of their neighborhoods, and resulting outcomes.

\end{itemize}

\section{Negotiation Protocol}
\subsection{Utility}

Let \textit{\( Nb_{t} \in N \textbackslash \{0\} \)} the set of the retailers in the neighborhood of each period \textit{t}, \textit{\(nb_{t}\)} the total quantity of retailers in \textit{\( Nb_{t}\)} and \textit{n} the total quantity of retailers in \textit{\( N_{C}\)}. The utility of each retailer \textit{c} is defined by the following equation:

\begin{equation} \label{fo2}
U_{c} = - ( 
\sum_{t \in T} (
h_{c}I_{ct}
+ \sum_{(i,j) \in A}\frac{(c_{ij}\sum_{k \in K}x_{ijkt})}{nb_{t}}
+ \frac{(up_{t} + fy_{t} + h_{0}I_{0t})}{n} ))
\end{equation}

The first term explains retailer \textit{c}'s specific inventory costs. The second term is the shipping price from the supplier, translated into transportation costs for neighborhoods where \textit{c} has deliveries and normalized by total retailers in those areas. The last term is the supplier's product price, translated into production, setup, and inventory costs for the entire horizon and normalized by the total quantity of retailers in the plan.

Retailers negotiate or vote in favor of a transaction only if their delta utility is positive, indicating lower costs and prices after changes. Notably, alterations to another retailer's delivery plan can affect the overall utility of that retailer.

\subsection{Agenda Transactions}

From retailers' delivery preferences, the supplier generates an optimal negotiation agenda and proposes changes to delivery plans, as following:

\begin{itemize}
\item Removal: The supplier identifies a chance to eliminate a retailer's delivery in a specific period, reducing shipping prices for that neighborhood. Note that if a removal happens, the same quantity must be added to another period where the retailer already has a delivery, adhering to demand and inventory constraints. This shouldn't trigger voting in the other neighborhood, as shipping prices aren't unit-specific.

\item Insertion: This happens when the supplier suggests adding a delivery for a retailer in a neighborhood where they had none before, increasing shipping prices for that area and reducing approval chances. The inserted quantity should be taken from another neighborhood with a current delivery to avoid triggering its voting phase. Insertions are accepted only if they significantly reduce product prices or when proposed along with removals.

\item Substitution: The supplier identifies an opportunity for an insertion and removal affecting two different neighborhoods. Voting in the first neighborhood is mostly against the change, while the second neighborhood is in favor. Since the first neighborhood outnumbers the second, the overall voting fails. To address this, the supplier proposes a substitution with another retailer's delivery in those neighborhoods to enhance acceptance chances.

\end{itemize}
Each proposal undergoes a transaction, negotiated among the retailers directly affected by changes to their delivery plans, and is subsequently voted on by retailers who are indirectly affected (experiencing an increase or decrease in shipping/product prices).

\begin{figure}[!ht]
  \centering
  \includegraphics[width=0.85\linewidth]{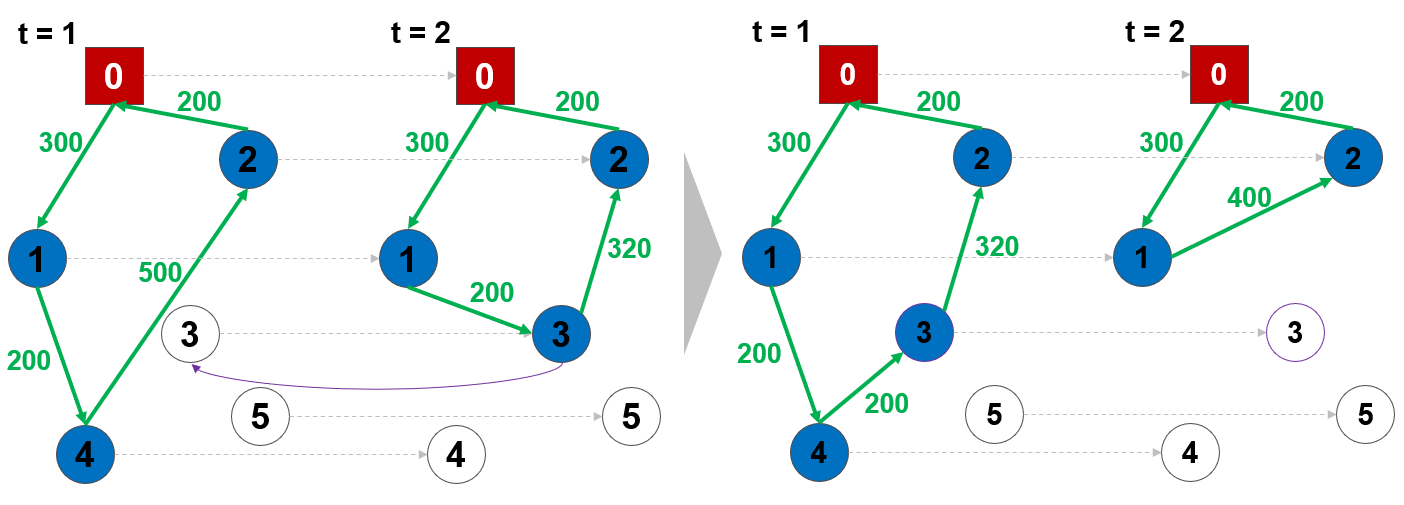}
  \caption{Transaction 1 - Insertion combined with a removal for a delivery of 10 units of product from retailer 3, anticipating it from period 2 to 1.}
  \label{fig:troca}
  \Description{Transaction 1 - Insertion combined with a removal for a delivery of 10 units of product from retailer 3, anticipating it from period 2 to 1.}
\end{figure}

\begin{figure}[!ht]
  \centering
  \includegraphics[width=0.85\linewidth]{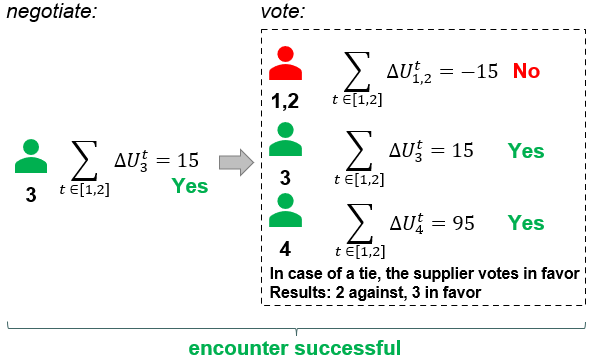}
  \caption{Transaction 1 - Negotiation and voting phases. As the delta utility from retailer 3 is positive for the proposed change, the voting phase is triggered. The result is a tie - 2 in favor and 2 against the change, so the supplier vote in favor in order to break the tie. This results in a successful transaction. (See complete calculation record in Appendix)}
  \label{fig:insertionwithremoval}
  \Description{Transaction 1 - Negotiation and voting phases. As the delta utility from retailer 3 is positive for the proposed change, the voting phase is triggered. The result is a tie: 2 in favor and 2 against the change, so the supplier vote in favor in order to break the tie. This results in a successful transaction. (See complete calculations in Appendix)}
\end{figure}

\begin{figure}[!ht]
  \centering
  \includegraphics[width=0.85\linewidth]{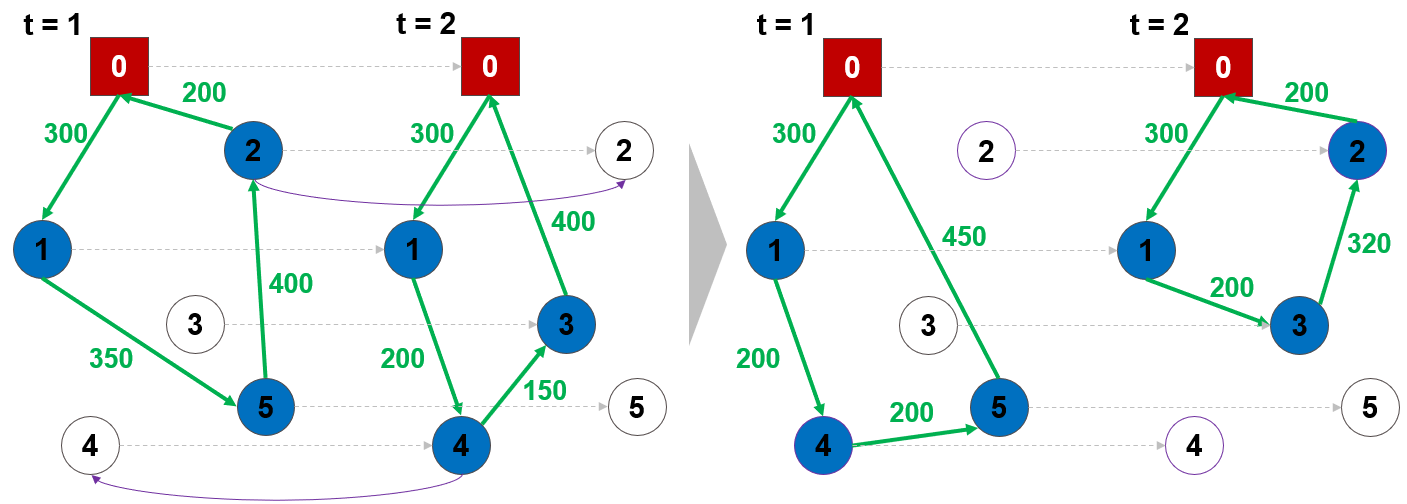}
  \caption{Transaction 2 - Substitution between two retailers' deliveries. Retailer 4 has a delivery of 5 units anticipated to period 1, while retailer 2 has a delivery of 15 units postponed to period 2.}
  \label{fig:substituicao}
  \Description{Transaction 2 - Substitution between two retailers' deliveries; retailer 4 has a delivery of 5 units anticipated to period 1, while retailer 2 has a delivery of 15 units postponed to period 2.}
\end{figure}

\begin{figure}[!ht]
  \centering
  \includegraphics[width=0.85\linewidth]{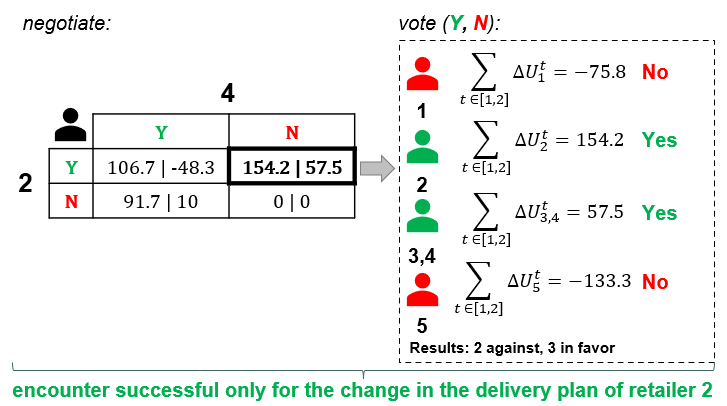}
  \caption{Transaction 2 - Negotiation and voting phases. Retailers 2 and 4 negotiate and decide  to apply changes only for retailer 2 (Y, N). Subsequently, the voting phase is triggered, with 3 votes in favor and 2 against. Then, the transaction succeeds only for the change in the delivery plan of retailer 2. (See complete calculation record in Appendix)}
  \label{fig:substitution}
  \Description{Transaction 2 - Substitution between two retailers' deliveries; retailer 4 has a delivery of 5 units anticipated to period 1, while retailer 2 has a delivery of 15 units postponed to period 2.}
\end{figure}

\subsection{Pseudo-Algorithm}

The full system flow is represented by the Algorithm 1.

\begin{algorithm}[!htbp] \label{algorithm1}
\caption{PRPPP}
\begin{algorithmic}[1]
\fontsize{7pt}{7pt}\selectfont
\State \textit{deliveryprefs}: retailers send supplier their delivery preferences
\State \textit{initialsol}: supplier generates a initial solution
\Procedure{ImprovePRPPP}{initial sol., delivery prefs.}
    \State Calculate stopping criterion
    \While{stopping criterion not met}
        \State \textit{optagenda}: supplier generates optimal agenda
        \For {transaction in agenda}
        \State \textit{negotiate}: ask the affected retailers to negotiate
        \State \textit{vote}: ask the affected neighborhoods to vote
        \If{\textit{negotiate} returns true and \textit{vote} returns true}
            \State Supplier updates the current delivery plan
        \Else
            \State Do not update the current delivery plan
        \EndIf  
        \EndFor
    \State Calculate stopping criterion
    \EndWhile
    \State \textbf{return} Final delivery plan
\EndProcedure
\end{algorithmic}
\end{algorithm}

\section{Conclusions and future work}

The study highlights MAS benefits in addressing critical challenges in applying optimization to supply chain planning, especially in the context of privacy preservation. The PRP, a known literature problem, is adapted to consider information constraints (PRPPP). This work proposes a hybrid MAS and optimization framework to solve it.

Future work involves defining algorithms for optimal agenda and initial solution generation, integrating heuristic algorithms. The establishment of a stopping criterion is necessary, considering that it should not be based only on a percentage of initial cost reduction, since the optimal solution is unknown. Furthermore, discrepancies between planning and execution may be addressed by adapting and automating the proposed framework for real-time optimization.

%%% The following command should be issued somewhere in the first column 
%%% of the final page of your paper.
\balance

%%%%%%%%%%%%%%%%%%%%%%%%%%%%%%%%%%%%%%%%%%%%%%%%%%%%%%%%%%%%%%%%%%%%%%%%

%%% The acknowledgments section is defined using the "acks" environment
%%% (rather than an unnumbered section). The use of this environment 
%%% ensures the proper identification of the section in the article 
%%% metadata, as well as the consistent spelling of the heading.

%%%%%%%%%%%%%%%%%%%%%%%%%%%%%%%%%%%%%%%%%%%%%%%%%%%%%%%%%%%%%%%%%%%%%%%%

%%% The next two lines define, first, the bibliography style to be 
%%% applied, and, second, the bibliography file to be used.

\bibliographystyle{ACM-Reference-Format}
\bibliography{references}

\newpage
\appendix

\section{APPENDIX- PRP Formulation}
Formulation by \cite{Boudia2007}:

Parameters:

\begin{itemize}
    \item \textit{u} unit production cost;
    \item \textit{f} fixed production setup cost;
    \item \textit{\(h_{i}\)} unit inventory cost at node \textit{i} (supplier and retailers);
    \item \textit{\(c_{ij}\)} transportation cost from node \textit{i} to node \textit{j};
    \item \textit{\(d_{it}\)} demand from retailer \textit{i} in period \textit{t};
    \item \textit{\(C\)} production capacity;
    \item \textit{\(Q\)} vehicle capacity;
    \item \textit{\(L_{i}\)}  maximum or target inventory level at node \textit{i};
    \item \textit{\(I_{i0}\)} initial inventory availabe at node \textit{i}.

\end{itemize}

Decision variables:

\begin{itemize}
    \item \textit{\(p_{t}\)} production quantity in period \textit{t};
    \item \textit{\(I_{it}\)} inventory at node \textit{i} at the end of period \textit{t};
    \item \textit{\(y_{t}\)} equal to 1 if there is production at the factory in period \textit{t}, 0 otherwise;
    \item \textit{\(z_{0kt}\)} equal to 1 if vehicle \textit{k} left the factory (node 0) in period \textit{t}, 0 otherwise;
    \item \textit{\(z_{ikt}\)} equal to 1 if customer \textit{i} was visited by vehicle \textit{k} in period \textit{t}, 0 otherwise;
    \item \textit{\(x_{ijkt}\)} equal to 1 if a vehicle travels directly from node \textit{i} to node \textit{j} in period \textit{t};
    \item \textit{\(q_{ikt}\)} quantity delivered to customer \textit{i} in period \textit{t}.

\end{itemize}

\begin{equation} \label{fo3}
min \sum_{t \in T}(up_{t} + fy_{t} + \sum_{i \in N}(h_{i}I_{it}) +\sum_{(i,j) \in A}(c_{ij}\sum_{k \in K}x_{ijkt} ))
\end{equation}

\begin{equation} \label{1}
s.t. \quad I_{0,t-1} + p_{t} = \sum_{i \in N_{C}}\sum_{k \in K} q_{ikt} + I_{0t} \quad \forall t \in T
\end{equation}

\begin{equation} \label{2}
I_{i, t-1} + \sum_{k \in K} q_{ikt} = d_{it} + I_{it} \quad \forall i \in N_{C}, \forall t \in T
\end{equation}

\begin{equation} \label{3}
p_{t} \leq M_{t}y_{t} \quad \forall t \in T
\end{equation}

\begin{equation} \label{4}
I_{0t} \leq L_{0} \quad \forall t \in T
\end{equation}

\begin{equation} \label{5}
I_{i,t-1}+\sum_{k \in K} q_{ikt} \leq L_{i} \quad \forall i \in N_{C}, \forall t \in T
\end{equation}

\begin{equation} \label{6}
q_{ikt} \leq M_{it} z_{ikt} \quad \forall k \in K, \forall i \in N_{C}, \forall t \in T
\end{equation}

\begin{equation} \label{7}
\sum_{k \in K}z_{ikt} \leq 1 \quad \forall i \in N_{C}, \forall t \in T
\end{equation}

\begin{equation} \label{8}
\sum_{j \in N}x_{jikt} + \sum_{j \in N}x_{ijkt} = 2z_{ikt} \quad \forall k \in K, \forall i \in N, \forall t \in T
\end{equation}

\begin{equation} \label{9}
\sum_{i \in N_{C}}q_{ikt} \leq Qz_{0kt} \quad \forall k \in K, \forall t \in T
\end{equation}

\begin{equation} \label{10}
\sum_{i \in S}\sum_{j \in S}x_{jikt} \leq |S| - 1 \quad \forall S \subseteq N_{C} : |S| \geq 2, \forall k \in K, \forall t \in T
\end{equation}

\begin{equation} \label{11}
p_{t}, I_{it}, q_{ikt} \geq 0 \quad \forall i \in N, \forall k \in K, \forall t \in T
\end{equation}

\begin{equation} \label{12}
y_{t}, z_{ikt}, x_{ijkt} \in {0,1} \quad \forall i,j \in N, \forall k \in K, \forall t \in T
\end{equation}

The objective function (\ref{fo3}) minimizes the total costs of production, production setup, factory and customer inventories, and delivery routing. Constraints (\ref{1})-(\ref{5}) represent the lot-sizing problem. Constraints (\ref{1}) and (\ref{2}) enforce the stock flow balance at the factory and customers, respectively. Constraint (\ref{3}) ensures that the production setup variable (\textit{\(y_{t}\)}) equals one if production occurs in a specific period and limits the production quantity to the minimum between the production capacity and the total demand in the remaining periods (\textit{\(M_{t}\)}). Constraints (\ref{4}) and (\ref{5}) restrict the maximum inventory at the factory and customers, respectively.

The remaining constraints, i.e., (\ref{6})-(\ref{10}), are the vehicle load and routing constraints. Constraints (\ref{6}) allow a positive delivery quantity only if customer \textit{i} is visited in period \textit{t}, and each customer can be visited by at most one vehicle (\ref{7}). Constraints (\ref{8}) ensure the flow conservation of vehicles. Constraints (\ref{9}) limit the quantity of product that can be transported by each vehicle. Constraints (\ref{10}) are the Subtour Elimination Constraints (SECs), similar to those in the Traveling Salesman Problem (TSP). Constraints (\ref{11}) and (\ref{12}) represent the domains of non-negative continuous variables and binary variables, respectively.

\section{Transactions' Calculation Record}

Assuming no change was made in the production plan, the retailer's utility every period \textit{t} is calculated as following:

\begin{equation}
U_{c}^{t} = (h_{c}I_{ct}) + \sum_{(i,j) \in A}(c_{ij}\sum_{k \in K}x_{ijkt})/nb_{t}
\end{equation}

The delta utility between two states of the same period \textit{t} can be calculated by subtracting the utility of the initial state from that of the final state:

\begin{equation}
\Delta U_{c}^{t} = U_{c, f}^{t} - U_{c, 0}^{t}
\end{equation}

Then, the total delta utility for two periods, e.g. \textit{t = 1} and \textit{t = 2}, can be calculated as follows:

\begin{equation}
\sum_{t \in [1, 2]}\Delta U_{c}^{t} = \Delta U_{c}^{1} + \Delta U_{c}^{2}
\end{equation}

The inventory cost \textit{\(h_{c}\)} is an intrinsic parameter from every retailer. The term \textit{\(I_{ct}\)} represents the decision variables of a retailer's demand plan. The shipping cost, denoted by \textit{\(\sum_{(i,j) \in A}(c_{ij}\sum_{k \in K}x_{ijkt})/nb_{t}\)}, illustrates the aggregate costs of each route segment highlighted in green in Figures \ref{fig:troca} and \ref{fig:substituicao} - the numerical values in green represent their calculated costs.

\subsection{Transaction 1}

\begin{equation}
\Delta U_{3}^{1} = U_{3, f}^{1} - U_{3, 0}^{1} = [- (2 * 0 + (200+320+200+200+300)/4)] - (0) = -325
\end{equation}

\begin{equation} 
\Delta U_{3}^{2} = U_{3, f}^{2} - U_{3, 0}^{2} = (0) - [-(2*10 + (200+320+200+300)/3)] = +340
\end{equation}

\begin{equation}
\begin{split}
\Delta U_{1,2,4}^{1} = - (200+320+200+200+300)/4)\\ - [-(200+500+200+300)/3]= 95
\end{split}
\end{equation}

\begin{equation}
\Delta U_{1,2}^{2} = - (200+400+300)/2) - [-(200+320+200+300)/3]= -110
\end{equation}

\begin{equation}
\Delta U_{4}^{2} = 0
\end{equation}

Thus,

\begin{equation}
\sum_{t \in [1, 2]}\Delta U_{3}^{t} = \Delta U_{3}^{1} + \Delta U_{3}^{2} = - 325 + 340 = 15
\end{equation}

\begin{equation}
\sum_{t \in [1, 2]}\Delta U_{1,2}^{t} = \Delta U_{1,2}^{1} + \Delta U_{1,2}^{2} = 95 + (- 110) = -15
\end{equation}

\begin{equation}
\sum_{t \in [1, 2]}\Delta U_{4}^{t} = \Delta U_{4}^{1} + \Delta U_{4}^{2} = 95 + 0 = 95
\end{equation}

\subsection{Transaction 2}

\begin{figure}[!ht]
  \centering
  \includegraphics[width=1\linewidth]{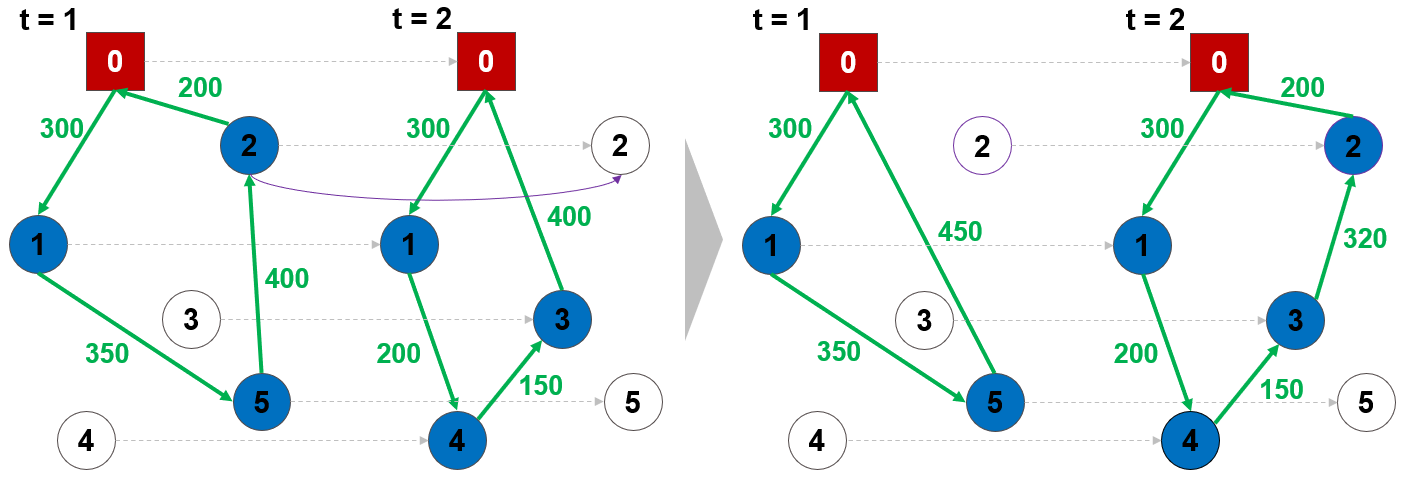}
  \caption{Transaction 2 - Substitution between two retailers' deliveries. Negotiation outcome (Y, N).}
  \label{fig:only2}
  \Description{Transaction 2 - Substitution between two retailers' deliveries. Negotiation outcome (Y, N).}
\end{figure}

\begin{figure}[!ht]
  \centering
  \includegraphics[width=1\linewidth]{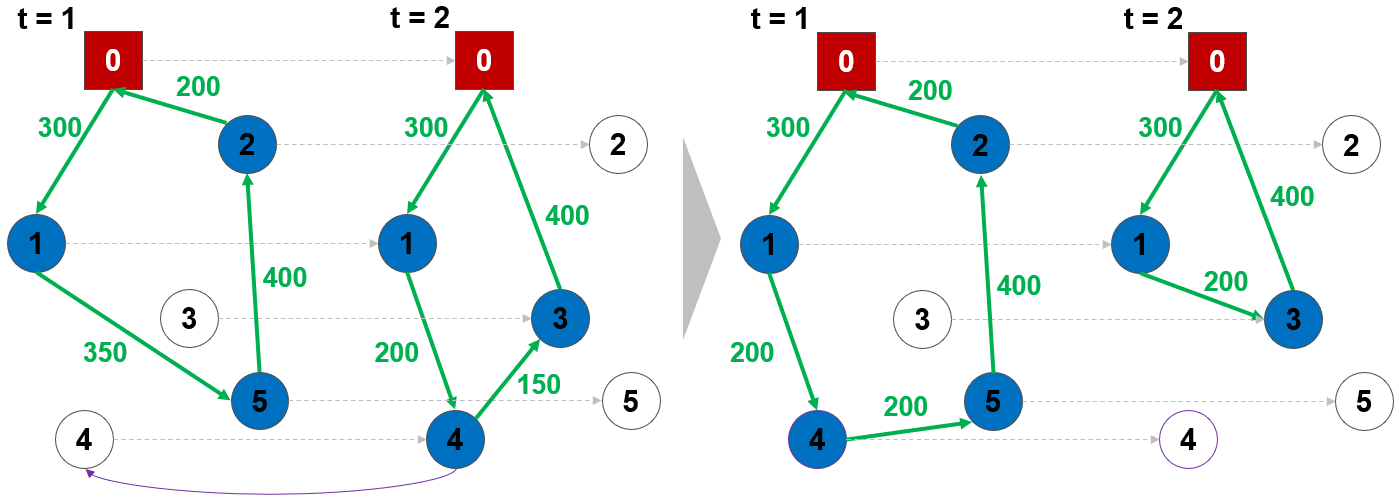}
  \caption{Transaction 2 - Substitution between two retailers' deliveries. Negotiation outcome (N, Y).}
  \label{fig:only4}
  \Description{Transaction 2 - Substitution between two retailers' deliveries. Negotiation outcome (N, Y).}
\end{figure}

Along with Figure \ref{fig:substituicao}, Figure \ref{fig:only2} and \ref{fig:only4} illustrates the Transaction 2 outcomes (Y, N) and (N, Y) respectively. Their updated shipping costs are represented in green. The total delta utility is calculated for every outcome.
For the (Y, Y) outcome:

\begin{equation}
\Delta U_{2}^{1} = U_{2, f}^{1} - U_{2, 0}^{1} = 0 - [- (2 * 15 + (200+400+350+300)/3)] = 446.7
\end{equation}

\begin{equation}
\Delta U_{2}^{2} = U_{2, f}^{2} - U_{2, 0}^{2} = [- (2 * 0 + (200+320+200+300)/3)] - (0) = -340
\end{equation}

\begin{equation}
\Delta U_{4}^{1} = - (3 * 5 + (450+200+200+300)/3) - (0) = -398.3
\end{equation}

\begin{equation}
\Delta U_{4}^{2} = 0 - [- (3 * 0 + (400+150+200+300)/3)] = 350
\end{equation}

Thus,

\begin{equation}
\sum_{t \in [1, 2]}\Delta U_{2}^{t} = \Delta U_{2}^{1} + \Delta U_{2}^{2} = 446.7 - 340 = 106.7
\end{equation}

\begin{equation}
\sum_{t \in [1, 2]}\Delta U_{4}^{t} = \Delta U_{4}^{1} + \Delta U_{4}^{2} = -398.3 + 350 = -48.3
\end{equation}

For the (Y, N) outcome:

\begin{equation}
\Delta U_{2}^{1} = 0 - [- (2 * 15 + (200+400+350+300)/3)] = 446.7
\end{equation}

\begin{equation}
\Delta U_{2}^{2} = - (2 * 0 + (200+320+150+200+300)/4)] - (0) = -292.5
\end{equation}

\begin{equation}
\Delta U_{4}^{1} = 0
\end{equation}

\begin{equation}
\begin{split}
\Delta U_{4}^{2} = - (3*0 + (200+320+150+200+300))/4\\ - [-(3*0 + (400+150+200+300)/3)] = 57.5
\end{split}
\end{equation}

Thus,

\begin{equation}
\sum_{t \in [1, 2]}\Delta U_{2}^{t} = \Delta U_{2}^{1} + \Delta U_{2}^{2} = 446.7 - 292.5 = 154.2
\end{equation}

\begin{equation}
\sum_{t \in [1, 2]}\Delta U_{4}^{t} = \Delta U_{4}^{1} + \Delta U_{4}^{2} = 0 + 57.8 = 57.5
\end{equation}

For the (N, Y) outcome:

\begin{equation}
\begin{split}
\Delta U_{2}^{1} = - [2*15 + (200+400+200+200+300)/4] \\ - [-(2*15 + (200+400+350+300)/3] = 91.7
\end{split}
\end{equation}

\begin{equation}
\Delta U_{2}^{2} = 0
\end{equation}

\begin{equation}
\Delta U_{4}^{1} = - (3 * 5 + (200+400+200+200+300)/4) - (0) = -340
\end{equation}

\begin{equation}
\Delta U_{4}^{2} = 0 - [- (3 * 0 + (400+150+200+300)/3)] = 350
\end{equation}

Thus,

\begin{equation}
\sum_{t \in [1, 2]}\Delta U_{2}^{t} = \Delta U_{2}^{1} + \Delta U_{2}^{2} = 91.7 + 0 = 91.7
\end{equation}

\begin{equation}
\sum_{t \in [1, 2]}\Delta U_{4}^{t} = \Delta U_{4}^{1} + \Delta U_{4}^{2} = -340 + 350 = 10
\end{equation}

Since the (Y, N) transaction outcome was chosen by retailers 2 and 4 in the negotiation, retailers 1, 3 and 5 are asked to vote in favor or against it. Thus, their delta utility for the voting phase is calculated as:

\begin{equation}
\Delta U_{1,5}^{1} = - (450+350+300)/2) - [-(200+400+350+300)/3] = -133.3
\end{equation}

\begin{equation}
\Delta U_{3}^{1} = 0
\end{equation}

\begin{equation}
\begin{split}
\Delta U_{1,3}^{2} = - (200+320+150+200+300)/4)\\ - [-(400+150+200+300)/3] = 57.5
\end{split}
\end{equation}

\begin{equation}
\Delta U_{5}^{2} = 0
\end{equation}

Thus,

\begin{equation}
\sum_{t \in [1, 2]}\Delta U_{1}^{t} = \Delta U_{1}^{1} + \Delta U_{1}^{2} = -133.3 + 57.5 = -75.8
\end{equation}

\begin{equation}
\sum_{t \in [1, 2]}\Delta U_{3}^{t} = \Delta U_{3}^{1} + \Delta U_{3}^{2} = 0+ 57.5 = 57.5
\end{equation}

\begin{equation}
\sum_{t \in [1, 2]}\Delta U_{5}^{t} = \Delta U_{5}^{1} + \Delta U_{5}^{2} = -133.3 + 0 = -133.3
\end{equation}

\end{document}